\documentclass[twocolumn,a4paper]{IEEEtran}
\usepackage{xcolor}
\usepackage[utf8]{inputenc}
\usepackage{amssymb}
\usepackage[intlimits]{amsmath}
\usepackage{amsfonts}
\usepackage{amsmath}
\usepackage{mathptmx}
\usepackage{graphicx}

\usepackage{listings}
\usepackage{algorithm}
\usepackage{algorithmicx}
\usepackage{algpseudocode}
\usepackage{tikz}
\usepackage{textcomp}
\usepackage[english]{babel}
\usepackage{bm}
\usepackage{xcolor}
\usepackage{booktabs} 

\newcommand{\colcit}[1]{\colorbox[rgb]{0.1,0.9,0.5}{\color[rgb]{0,0,0}#1}}

\renewcommand{\colcit}[1]{#1}

\newcommand\copyrighttext{%
  \footnotesize  This work has been submitted to the IEEE for possible publication. Copyright may be transferred without notice, after which this version may no longer be accessible. }
\newcommand\copyrightnotice{%
\begin{tikzpicture}[remember picture,overlay]
\node[anchor=south,yshift=20pt] at (current page.south) {\fbox{\parbox{\dimexpr\textwidth-\fboxsep-\fboxrule\relax}{\copyrighttext}}};
\end{tikzpicture}%
}

\begin{document}
\title{Analytical Interpretation of Latent Codes in InfoGAN with SAR Images }
	\author{Zhenpeng Feng, \textit{Student Member, IEEE,} Milo\v s Dakovi\' c, \textit{Member, IEEE,}
	Hongbing Ji, \textit{Senior Member, IEEE,} Mingzhe Zhu,  Ljubi\v sa Stankovi\'c, \textit{Fellow, IEEE} 
\thanks{ Z. Feng, H. Ji and M. Zhu are with School of Electronic Engineering, Xidian University, Xi'an, China. Email: zpfeng\_1@stu.xidian.edu.cn,  zhumz@mail.xidian.edu.cn, hbji@mail.xidian.edu.cn. \newline \indent M. Dakovi\' c and L. Stankovi\' c are with the EE Department of the University of Montenegro, Podgorica, Montenegro. Email: \{ljubisa,  milos\}@ucg.ac.me.}

}

	\maketitle
\copyrightnotice

\begin{abstract}
Generative Adversarial Networks (GANs) can synthesize abundant photo-realistic synthetic aperture radar (SAR) images. Some recent GANs (e.g., InfoGAN), are even able to edit specific properties of the synthesized images by introducing latent codes. It is crucial for SAR image synthesis since the targets in real SAR images are with different properties due to the imaging mechanism. Despite the success of InfoGAN in manipulating properties, there still lacks a clear explanation of how these latent codes affect synthesized properties, thus editing specific properties usually relies on empirical trials, unreliable and time-consuming. In this paper, we show that latent codes are disentangled to affect the properties of SAR images in a non-linear manner. By introducing some property estimators for latent codes, we are able to provide a completely analytical nonlinear model to decompose the entangled causality between latent codes and different properties. The qualitative and quantitative experimental results further reveal that the properties can be calculated by latent codes, inversely, the satisfying latent codes can be estimated given desired properties. In this case, properties can be manipulated by latent codes as we expect.
	\end{abstract}

\section{Introduction}

Synthetic aperture radar (SAR) is considered a well-established technology for providing day-and-night and weather-independent images, widely used in geological exploration, ocean research, disaster monitoring, military, environmental, and earth system monitoring, etc.  \colcit{\cite{ender2014recent, moreira2013tutorial,VehicleborneSARImaging,GroundMovingTarget,berizzi2018radar,vesna}}. However, SAR is always an expensive means of monitoring because the expenditure of airplane flights or launching satellites is much higher than other optical or infrared imaging devices \colcit{\cite{EfficientSimulationofhybrid,DataAugmentation}}. Therefore, the cost of obtaining abundant SAR images is quite high.

To obtain such SAR images in an efficient, effective, and economic manner, numerous generative models are utilized to synthesize SAR images and one of the most promising is Generative Adversarial Network (GAN) \colcit{\cite{GANBasedSARtoOpticalImage,SARImagesGeneration,PSGAN,HPGAN}}. GAN is proposed by Goodfellow. et al., containing a generator network, $G$, and a discriminator network, $D$ \colcit{\cite{GAN1,GAN2}}. The generator manages to approximate the real data distribution from a random noise distribution, and the discriminator estimates the probability that the input sample is a real image or synthesized by the generator. Such optimization is achieved by a minimax two-player game, thus it is termed "adversarial". It should be noted that GAN only adopts a simple noise vector as the input to $G$ without imposing any restrictions on how the generator uses this noise \colcit{\cite{GAN2}}. In this case, the direction of image generation can be hardly controlled as we expect since the noise is used by the generator in a highly entangled way \colcit{\cite{Semantichierarchy}}. However, SAR images naturally include some semantically meaningful properties due to the imaging mechanism. For instance, some rotation, translation, and scaling of the target usually emerge with different view angles between radar and the target \colcit{\cite{GANBasedSARtoOpticalImage}}. To further control the generation direction of GAN, X. Chen, et al. proposed InfoGAN to further disentangle the input noise by introducing latent codes \colcit{\cite{InfoGAN}}. A strong correlation between latent codes and those properties will be established by maximizing their mutual information during InfoGAN's training. 

Although InfoGAN can generate SAR images with semantically meaningful properties by latent codes, the relation between properties and latent codes still lacks clear analytical interpretation \colcit{\cite{Semantichierarchy,InfoGANSAR}}. It brings in two problems: (1) How to obtain the property value from latent codes? (2) How to obtain satisfying latent codes when a desired property value is given? Obviously, they are not easy to solve in InfoGAN. In this paper, various property estimators are introduced to measure such relation. The results show that a single latent code retains an approximately $\tanh$ relation with a certain property while multiple latent codes are entangled to edit different properties in a complex nonlinear manner. The main contributions of this paper is that a completely analytical relation is provided between latent codes and properties, providing possibility to  edit the properties by manipulating latent codes as we expect.

The rest of this paper is organized as follows. Section \ref{sec:Background} introduces how these properties emerge in SAR imaging and the mechanism of InfoGAN. Section \ref{sec:Methodology} describes how to quantify the relation between properties and latent codes.  In Section \ref{sec:Experiment}, experimental results with fully-simulated, semi-simulated, real SAR images (with/without background) in various cases will be provided and analyzed. Section \ref{sec:Conclusion}  concludes this paper.
\section{Background Knowledge and Motivation}\label{sec:Background}
\subsection{Basic SAR Principles}

\begin{figure*}[tbp]
	\centering
	\includegraphics[]{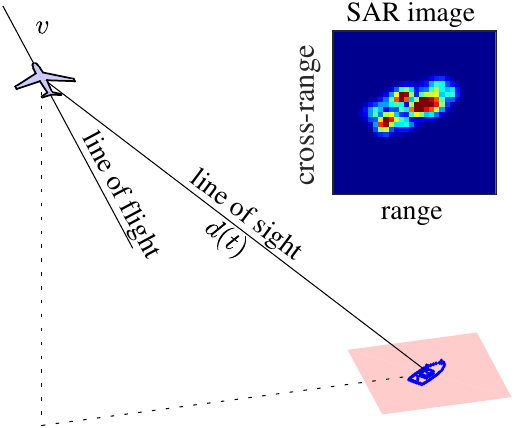} \hspace{0mm}
	\includegraphics[]{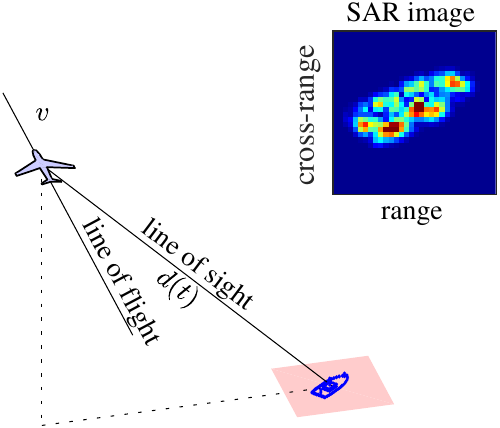} \hspace{0mm}
	\includegraphics[]{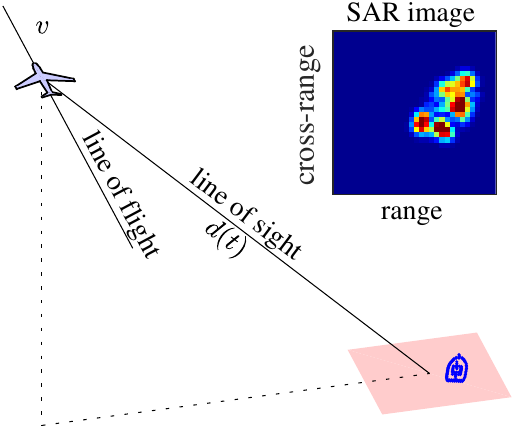}
	\caption{Sythetic aparature radar (SAR) setup with various relative positions of the radar and the target. The mechanism of SAR imaging (left). The emergence of scaling of the target in a SAR image (middle). The emergence of rotation and translation of the target in a SAR image (right)}
	\label{propertiesofSAR}
\end{figure*}

A radar image is obtained by transmitting repeated pulses and processing the echoes returned from the target \colcit{\cite{Martorella2,Wu, Narayanan1,LJS2,LJS,Martorella3,Narayana2}}. A common choice for the pulse is a linear frequency-modulated continuous-wave (LFM-CW) signal, transmitted in a form of a series of chirps. The received signal, which is scattered from a target, is delayed and changed in amplitude as compared to the transmitted signal, containing in that way the information about the target position and reflectivity . The received signal from an elementary (a point) scatterer, after an appropriate mixing with the transmitted signal, demodulation, compensation, and residual video phase  filtering, is of the form\cite{ender2014recent} 
\begin{align}
	S(m,t)	= \sigma \exp \left( j\omega_{0} \tfrac{2d(t)}{c} \right) \exp\left(-j2\pi \tfrac{B(t-mT_r)}{T_r}\tfrac{2d(t)}{c}\right)
\end{align}
where $\sigma$ is the reflection coefficient of the scattering point, $\omega_{0}$ is the radar operating frequency, $\exp(j\omega_{0} \frac{2d(t)}{c})$ is the scattering phase and $\exp(-j2\pi \frac{B}{T_r}(t-mT_r)\frac{2d(t)}{c})$ describes 
the phase variation due to the varing distance. The transmission and receiving procedure is repeated every $T_r$ seconds (the pulse
repetition interval - PRI).

In SAR images the radar platform movement is crucial in producing a high-resolution image.  Therefore, the SAR systems are based on a pulsed radar installed on a platform with a forward movement.  The distance between the radar moving at constant velocity $v$ and a point target on the ground can be described as \colcit{\cite{moreira2013tutorial}}
\begin{align}
	d(t) = \sqrt{d_0^{2}+(vt)^{2}}
\end{align}
where $t=0$ is the time of closest approach, when the distance is minimum as $d(0)=d_0$.
Assume $M$ pluses are transmitted and $N$ range cells are inside a pulse interval, $t=n T_s$. The received echo signal can form a $M\times N$
data matrix of complex samples. The column dimension
corresponds to the range direction. Note the radar acquires a range line in each PRI thus forming the row dimension
of the data matrix, termed azimuth direction. In the case of multi-point targets, the superposition principle applies.
Therefore, the raw SAR data are the echoes from the
illuminated scene (of multiple points or even continuous targets) sampled both in range direction and azimuth direction.

Different from optical sensors, however, raw SAR data does
not provide any visible information on the scene \colcit{\cite{ender2014recent}}. It is only after basic SAR processing steps that an image is obtained. In a very simplified way, the complete processing can be understood as two separate matched filter operations along with the range and azimuth dimensions, instead of performing a convolution in the time domain, multiplication in the frequency domain
is adopted due to the much lower computational load. The first step is to compress the transmitted chirp signals to
a short pulse.  Azimuth compression follows the same basic reasoning, that is, the signal is convolved with its reference function, which is the complex conjugate of the response expected from a point target on the ground. The SAR image is efficiently calculated using, for example, the two-dimensional fast Fourier transform (FFT) codes  \cite{53462}. 


To know a target or scene for analysis, detection, or classification, it is desirable to have its SAR image acquired from different positions \colcit{\cite{CNN,ExplainabilityofDeepSAR}}. 
Different relative viewing angles  (resulting from changes of flight direction or target movement in different revisits) results in a kind of target rotation in SAR image. The radar revisits could be also conducted from different distances to the target or the target could move between revisits resulting in a kind of target shifting and/or scaling in SAR image. These kinds of target changes in radar image will be referred as properties of the target, as illustrated in Fig.~\ref{propertiesofSAR}.
In some cases, numerous revisits or observations may be expensive or in some hostile or unique environments even not possible. Then it would be of interest to use the available set of data and try to synthesize new possible images, preferably with controlled properties, defined by, for example, different rotations, translation, and scaling that would at the same time fully correspond to the existing data. To this aim, we will present and apply GAN and InfoGAN.

\subsection{GAN and InfoGAN}

\begin{figure*}[tbp]
	\centering
	\includegraphics[width=\textwidth]{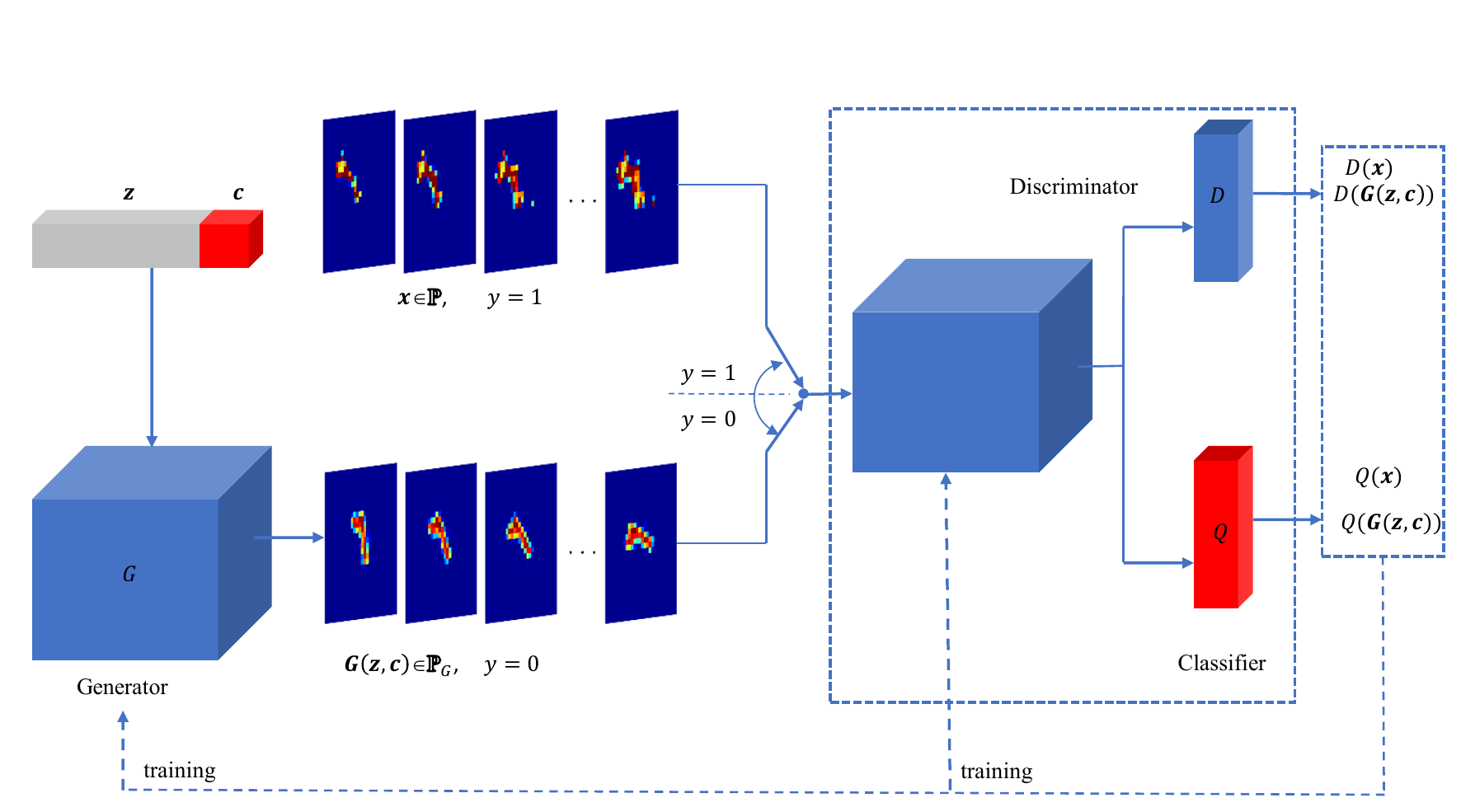}
	\caption{The architecture of GAN and InfoGAN. The basic GAN is obtained by excluding the red blocks and latent codes $\mathbf{c}$.}
	\label{infoGAN}
\end{figure*}

The main task of generative adversarial network is to train a transposed neural network to produce images that match real images $\mathbf{x}_n$ from a set $\mathbb{P}$ \colcit{\cite{GAN1,BrainComputerInterface}}.
It means that GAN learns a generator (transposed convolution neural network), denoted by G, to synthesize images as close to $\mathbb{P}$ by feeding the generator with a noise vector $\mathbf{z}$, commonly Gaussian or uniformly distributed. $\mathbf{G}\mathbf{(z)}$ denotes an image from a set of generated images, $\mathbb{P}_G$. The generator is trained against an adversarial discriminator network, D,  whose structure corresponds to a convolution neural network with the aim to distinguish (discriminate the cases) if the sample image as the input to the discriminator is from the true data set of images,   $\mathbb{P}$, or from the generator produced set of images.  $\mathbb{P}_G$. The basic structure of a GAN is included in Fig.~\ref{infoGAN}.

After both networks, the generator and the discriminator, are initialized by random weights, the training process is defined based on the loss function. First, we will consider the discriminator only. 
At its input, we have an image (as it is common for a convolutional neural network), either a sample image $\mathbf{x}$ from the set of real data, $\mathbb{P}$, or a synthesized image, $\mathbf{G}\mathbf{(z)}$, produced by the generator with a random input noise, $\mathbf{z}$. 
The output of the discriminator is a scalar denoted by $D( \cdot)$. It is either $D(\mathbf{x})$ or $D(\mathbf{G}\mathbf{(z)})$. The output value of the discriminator is a normalized such that $0 \le D(\mathbf{x}), \  D(\mathbf{G}\mathbf{(z)}) \le 1$. 
The aim of the discriminator is to  discriminate the cases when the input is: (i) a real image from $\mathbb{P}(\mathbf{x})$ or (ii) a generated ``fake'' (synthesized) image $\mathbf{G}\mathbf{(z)}$, by learning to produce the 
output values $D(\mathbf{x})$ close to $1$ and the values  $D(\mathbf{G}\mathbf{(z)})$ close to $0$. The target signal, which will be used during the supervised learning, will be denoted by $y_{\mathbf{x}}$. It assumes that the values:
\begin{enumerate}
 \item $y_{\mathbf{x}}=1$ if the input to the discriminator is a real image $\mathbf{x}$ from the set $\mathbb{P}(\mathbf{x})$; 
 \item $y_{\mathbf{x}}=0$ if the input to the discriminator is a synthesized image, $G\mathbf{(z)}$, being output from the generator. 
\end{enumerate} 
  The value of the target signal, $y_{\mathbf{x}}$, is provided at the output of the discriminator as reference signal for the loss function calculation during the training process.  A simple loss function could be in a quadratic form
  \begin{equation}
  	\mathcal{L}(D)=y_{\mathbf{x}}D^2(\mathbf{x})+ (1-y_{\mathbf{x}})(1-D(\mathbf{G}(\mathbf{z})))^2.
  \end{equation}
 This function assumes only one of two values $\mathcal{L} \in \{ D^2(\mathbf{x}),(1-D(\mathbf{G}(\mathbf{z}))^2 \}$. Since  $0 \le D(\mathbf{x})$, $ D(G\mathbf{(z)}) \le 1$, the loss function will reach its maximum value $\mathcal{L}(D)=1$ for any input to the discriminator, either $\mathbf{x}$ of $\mathbf{G}(\mathbf{x})$, if $D(\mathbf{x})=1$ and $D(G\mathbf{(z)})=0$. Therefore, by maximizing the loss function $\mathcal{L}(D)$, we can achieve the ideal discriminator performance. 
  
 In the GAN, the cross-entropy form of the loss function is used (with the same aim and the same qualitative analysis as in the quadratic loss function) \colcit{\cite{ElectromagneticInverseScattering}}. The cross-entropy form of the loss function is defined by $y_{\mathbf{x}}\log D(\mathbf{x})+ (1-y_{\mathbf{x}})\log(1-D(\mathbf{G}(\mathbf{z}))$, with the learning process for the discriminator neural network defined as
  \begin{equation}
  \max_D\mathcal{L}(D)=\max_D\{y_{\mathbf{x}}\log D(\mathbf{x})+ (1-y_{\mathbf{x}})\log(1-D(\mathbf{G}(\mathbf{z}))\}.
  \end{equation}
It is easy to conclude that, for $0 \le D(\mathbf{x}), D(G\mathbf{(z)}) \le 1$, this loss function achieves its maximum $\mathcal{L}(D)=0$ when $D(\mathbf{x})=1$ and $D(G\mathbf{(z)})=0$. 

Maximization of the cross-entropy loss function is commonly done over a set (mini-batch) of input real images, $\mathbf{x}_1$, $\mathbf{x}_2$, ..., $\mathbf{x}_m$, and generated images  $\mathbf{G}(\mathbf{z}_1), \mathbf{G}(\mathbf{z}_2),\ldots , \mathbf{G}(\mathbf{z}_m)$. The expression for the cross-entropy loss function will be also simplified by omitting $y_{\mathbf{x}_i}$.  Namely, it will be assumed that the input to discriminator is fed by alternating 
$\mathbf{x}_1$ and $\mathbf{G}(\mathbf{z}_1)$, then  $\mathbf{x}_2$ and $\mathbf{G}(\mathbf{z}_2)$, and so on in succession until $\mathbf{x}_m$ and $\mathbf{G}(\mathbf{z}_m)$. In this way, we may write two loss function terms: (i) $\log D(\mathbf{x}_i)$ for  $\mathbf{x}_i$ and (ii) $\log(1-D(\mathbf{G}(\mathbf{z}_i)))$ for  $\mathbf{G}(\mathbf{z}_i)$   as $\log D(\mathbf{x}_i)+ \log(1-D(\mathbf{G}(\mathbf{z}_i)))$, for each $i=1,2,\dots,m$. The mean value over $2m$ images ($m$ real images and $m$ generated images) is then defined by    
\begin{equation}
\max_D\mathcal{L}(D)=\max_D\frac{1}{m}\sum_{i=1}^m \left( \log D(\mathbf{x}_i)+ \log(1-D(\mathbf{G}(\mathbf{z}_i)))\right). \label{lLF}
\end{equation}

After the discriminator is trained (in the first cycle) based on the loss function (\ref{lLF}), its weights are frozen and the generator network is now trained for this cycle.  Since the generator does not have any knowledge about the real images, the part $\log D(\mathbf{x})$ is not used in the loss function for the generator weight training (only generated images are used, when  $y_{\mathbf{x}_i}=0$). The aim of the generator is to produce images as similar to those from the set $\mathbb{P}(\mathbf{x})$ as possible. Within the loss function framework, this aim will be achieved if the generator can close the gap between the discriminator output values $D(\mathbf{x})$ and $D(\mathbf{G}(\mathbf{x}))$ as much as possible. Since it can not change $D(\mathbf{x})$, this should be done by increasing the value of $D(\mathbf{G}(\mathbf{x}))$ toward $1$ or, in other words, by making the new loss function $\mathcal{L}(G)=\log(1-D(\mathbf{G}(\mathbf{z})))$ as small as possible, that is (within the same mini-batch), find
\begin{equation}
	\min_G\{\frac{1}{m}\sum_{i=1}^m \log(1-D(\mathbf{G}(\mathbf{z}_i)))\}.
\end{equation}

After the generator is trained in this way (in the first cycle), its weights are frozen and the discriminator  network is trained again within the second cycle. These cycles are continued for a defined number of echoes, when the GAN is assumed to be fully trained.  In the ideal case, after the training is finished, the discriminator will not be able to discriminate the real and the synthesized images from generator, meaning it will produce the output $D(\mathbf{x})=D(G\mathbf{(z)})=1/2$ and the loss function value of form (\ref{lLF}) will be $\mathcal{L}(D)=2 \log(1/2)=-4$. 

The combined loss function of GAN for both the discriminator and the generator can be summarized by the following expression:
\begin{align}
	\min\limits_{G} \max\limits_{D} \mathcal{L}(G,D)  = {} & \mathbb{E}_{\mathbf{x}} \{\log D(\mathbf{x}) \} \label{combGan}  + \mathbb{E}_{\mathbf{z}}\{\log(1-D(\mathbf{G}(\mathbf{z})))\}. 
\end{align}

It is clear from (\ref{combGan}) that no restrictions are imposed on the input noise data, thus the properties are highly entangled in generated images. To generate images with semantically meaningful properties, InfoGAN introduces latent codes, $\mathbf{c}=[c_1,c_2,\dots,c_n]$, and a classifier, Q, with the same architecture sharing the trainable parameters with discriminator. The purpose of classifier is to maximize the mutual information $I(\mathbf{c};\mathbf{G}(\mathbf{z},\mathbf{c}))$ between $\mathbf{c}$ and $\mathbf{G}(\mathbf{z},\mathbf{c})$, defined as:
\begin{align}
I(\mathbf{c};\mathbf{G}(\mathbf{z},\mathbf{c}))= H(\mathbf{c}) - H(\mathbf{c}|(\mathbf{z},\mathbf{c}))
\end{align}
where $H(\mathbf{c}) = -\sum_{i} p(c_{i})\log(p(c_{i}))$ is the entropy of $\mathbf{c}=[c_1,c_2,\dots,c_n]$. The mutual information $I(\mathbf{c};\mathbf{G}(\mathbf{z},\mathbf{c}))$ means that if $\mathbf{c}$ and $\mathbf{G}(\mathbf{z},\mathbf{c})$ are independent, then $I(\mathbf{c};\mathbf{G}(\mathbf{z},\mathbf{c}))=0$, because knowing $c$ reveals
nothing about the  $\mathbf{G}(\mathbf{z},\mathbf{c})$ (degrade to classic GAN); by contrast, if $\mathbf{c}$ and $\mathbf{G}(\mathbf{z},\mathbf{c})$ are strongly related, then maximal mutual information is attained. It means that the information in the latent code $\mathbf{c}$ should not be lost in the generation process. Hence, the information-regularized loss function is as follows:
\begin{align}
	\min\limits_{G} \max\limits_{D} \mathcal{L}_I(G,D)={} & \mathbb{E}_{\mathbf{x}} \{\log D(\mathbf{x}) \}+ \mathbb{E}_{\mathbf{z}}\{\log(1-D(\mathbf{G}(\mathbf{z})))\} \\
 + \lambda I(\mathbf{c};\mathbf{G}(\mathbf{z},\mathbf{c})).
\end{align}
Fig.~\ref{infoGAN} shows the architecture of an InfoGAN.

\section{Methodology}\label{sec:Methodology}

Next we will consider SAR images of the target taken with various setups and relate them to the latent codes in InfoGAN. The aim is to train InfoGAN to synthesize available images with various target properties and to produce new ones by changing latent codes. This process could be controlled by relating the latent codes to the SAR image transformations. Cases with one and two properties will be considered. In the analysis of one property we will use one or two latent codes, while in the case of two-properties two latent codes are used. 
  

\subsection{Property measurement}

When the radar illuminates a target (for example, a vehicle, a ship, or any other object of interest) in two different visits, SAR images may differ due to different viewing angles, target maneuvering, or different distance between the radar and the target in these two illuminations. The changes in radar image can be described by a rotation (with possible changes in the reflectivity or visibility of some scatterers in the target). Other possible change in the SAR image results from the possible distance change between the radar and the target, and may be described by a scaling of the target in SAR image (with possible changes in the radar image structure due to the fusing or separation of close scatterers due to the resolution values). This will be referred as the scaling property. Also, the target relative position can be changed in two different illuminations, causing the shifts in the radar image.  

To quantify these properties of radar images, we should introduce their relative measures with respect to one SAR image, assumed to be the reference image. To this aim, we will use the cross-correlation function to evaluate the similarity between two images \colcit{\cite{Correlation}}. Assume $\mathbf{X}$ and $\mathbf{Y}$ are two images of the same size, $N\times N$. The cross-correlation between these two images,  $r(\mathbf{X},\mathbf{Y})$,is defined as
\begin{align}\label{eq:res1}
	r(\mathbf{X},\mathbf{Y}) &= \frac{\sum_{i}\sum_{j}(X(i,j)-\bar{X})\sum_{i}\sum_{j}(Y(i,j)-\bar{Y})}{\sqrt{\sum_{i}\sum_{j}(X(i,j)-\bar{X})^2} \sqrt{\sum_{i}\sum_{j}(Y(i,j)-\bar{Y})^2}}\\
	\bar{X} &= \frac{1}{N^2}\sum_{i}\sum_{j}X(i,j), \qquad \bar{Y} = \frac{1}{N^2}\sum_{i}\sum_{j}Y(i,j)  \label{eq:res1b}
\end{align}
where $\bar{X}$ and  $\bar{Y}$ denote the mean of images $\mathbf{X}$ and $\mathbf{Y}$, and the denominator normalizes the cross-correlation to the range from $0$ to $1$. The summation range is from $1$ to $N$ for all sums in (\ref{eq:res1}) and (\ref{eq:res1b}). It can be observed that $r(\mathbf{X},\mathbf{Y})$ will be $1$ if $\mathbf{X}=\mathbf{Y}$, and $r(\mathbf{X},\mathbf{Y})$ will assume value smaller than 1 if $\mathbf{X}$ is becoming more different from $\mathbf{Y}$. 

If we want to use cross-correlation to measure the translation of a target $\mathbf{I}_j$ with respect to the reference image $\mathbf{I}_0$ then we will perform the translation operation of the reference image $\mathbf{I}_0$ for different $d_x$ with steps $\Delta d_x$ and $d_y$ with steps $\Delta d_y$, denoted by $\mathcal{T}_{\delta}\{\mathbf{I}_0\}$, and find the resulting translation parameter as the position $d_x$, $d_y$ when the maximum of the function $r(\mathcal{T}_{\delta}\{\mathbf{I}_0\},\mathbf{I}_j)$  is found
\begin{align}\label{eq:transition}
	\boldsymbol{\delta}_S(j)=\arg \max_{\boldsymbol{\delta}} \{r(\mathcal{T}_{\delta}\{\mathbf{I}_0\},\mathbf{I}_j)\},  
\end{align}  
where $\boldsymbol{\delta}_S$ is, in general, a vector, with corresponding shifts in the direction of range and cross-range \cite{vesna}.  

In a similar way, we say that the original image is rotated for $\delta_R$ when the maximum of the cross-correlation between the reference image, rotated for an angle $\delta_R$, and the considered image $\mathbf{I}_j$, is found, that is 
\begin{align}\label{eq:rotation}
	\delta_R(j)=\arg \max_{\delta} \{r(\mathcal{R}_{\delta}\{\mathbf{I}_0\},\mathbf{I}_j)\},  
\end{align}  
where now $\mathcal{R}_{\delta}\{\mathbf{I}_0\}$ denotes the reference image rotated for an angle $\delta_R(j)$. The rotated and reference image may differ in reflectivity, meaning that the maximum value of the cross-correlation will not be equal to one. In order to reduce the influence of the variations in the reflectivity during the rotations, we can introduce thresholding (limiting) or even consider only the support functions (support function of an image assumes value $0$ where the image is $0$ or close to $0$ and $1$ otherwise) of the considered objects. The rotation parameter is then calculated as 
\begin{align}\label{eq:rotationTH}
\delta_R(j)=\arg \max_{\delta} \{r(\mathcal{R}_{\delta}\{\mathbf{H}_T\{\mathbf{I}_0\}\},\mathbf{H}_T\{\mathbf{I}_j\})\},
\end{align}
 where $\mathbf{H}_T\{\mathbf{I}\}$ denotes the limited  version of the image $\mathbf{I}$, with a threshold $T$, that is
\begin{align}\label{eq:rotationTHDef}
\mathbf{H}_T\{I(i,j)\} =\begin{cases} I(i,j) &\text{for } \ I(i,j) \le T \\
T &\text{for } \ I(i,j) > T. \end{cases}
\end{align}

Finally the scaling property is defined in the same way, as the position of the maximum of correlation between the considered image $\mathbf{I}_j$ and the scaled reference image $\mathcal{S}_{\delta}\{\mathbf{I}_0\}$ for a scaling parameter $\delta$, that is 
\begin{align}\label{eq:scaling}
	\delta_A(j)=\arg \max_{\delta} \{r(\mathcal{S}_{\delta}\{\mathbf{I}_0\},\mathbf{I}_j)\}. 
\end{align} 

After we introduced measures of various mage transformations, we are now ready to relate them with latent codes in the InfoGAN.
%
%
\subsection{Relation of the properties and latent codes}

\noindent\textbf{ One property - One latent code:} Next we assume that the InfoGAN is trained with $P$ real SAR images when one of the considered properties (for example, relative angle of the target with respect to the radar direction) changes.   	
After the learning process, the InfoGAN is able to synthesize the corresponding SAR images, in an ideal case the same as the real original images, with the latent code $\mathbf{c}_1$, being related to the property change in the particular SAR images. After the learning process has finished, we generate a new set of $K$ latent code values $\mathbf{c}_1=[c_1(1), c_1(2), \dots, c_1(K)]^T$. Then, a set of images is generated using the values $c_1(k)$, $k=1,2,\dots,K$ and random input noises $\mathbf{z}_k$. The obtained images are denoted by 
\begin{align}\label{eq:scalinggenerImag}
	\mathbf{I}_k=\mathbf{G}(\mathbf{z}_k,c_1(k)), \ \ \ k=1,2,\dots,K. 
\end{align} 
 Then we use one of measures (\ref{eq:transition}), (\ref{eq:rotationTH}), or (\ref{eq:scaling})  to calculate the measure of properties for the each synthesized SAR image from the set.  The relative measure of the rotation with respect to the reference image $\mathbf{I}_0$ is calculated using
 \begin{align}\label{eq:rotation_corr}
 	\begin{split}
 		\delta_{R}(1) &= \arg \max \{r(\mathcal{R}_{\delta}\{\mathbf{H}_T\{\mathbf{I}_0\}\},\mathbf{H}_T\{\mathbf{I}_1\})\} \\
 		\delta_{R}(2)& = \arg \max \{r(\mathcal{R}_{\delta}\{\mathbf{H}_T\{\mathbf{I}_0\}\},\mathbf{H}_T\{\mathbf{I}_2\})\}\\
 		&\cdots\\
 		\delta_{R}(K) &= \arg \max \{r(\mathcal{R}_{\delta}\{\mathbf{H}_T\{\mathbf{I}_0\}\},\mathbf{H}_T\{\mathbf{I}_K\})\}
 	\end{split} 
 \end{align} 
 
(a) \textit{Linear model}: For the rough analysis, we consider a linear model for the approximation of the obtained measure of rotation and the latent code used to produce the corresponding image 
\begin{align}\label{eq:rotationLinear11}
	\hat{\delta}_R(k) = v_{1}c_1(k) + v_{0}, , \ \ \ k=1,2,\dots,K.
\end{align}
where $v_0$ and $v_1$ are two unknown parameters. To estimate them, we can write a matrix form of these equations 
\begin{align}\label{eq:rotation_linear_K_matrix}
	\hat{\boldsymbol{\delta}}_{R}=
	\begin{bmatrix}\hat{\delta}_{R}(1) \\ 
		\hat{\delta}_{R}(2) \\
		\vdots\\
		\hat{\delta}_{R}(K) \\	
	\end{bmatrix}
	=	
	\begin{bmatrix} c_{1}(1) & 1 \\ 
		c_{1}(2) & 1\\
		\vdots\\
		c_{1}(K) & 1\\	
	\end{bmatrix} \begin{bmatrix}v_1 \\ v_0	\end{bmatrix}=\mathbf{A}\mathbf{V},
\end{align}
 where $\mathbf{A}$ is matrix with a latent codes column and a column with elements equal to $1$, and $\mathbf{V}=	[v_1, \ v_0]^T$.

 Now we can obtain the optimal parameters $v_0$ and $v_1$ by optimizing the following equation:
\begin{align}\label{eq: optimize_v}
 \mathbf{V}= \arg \min \|\boldsymbol{\delta}_{R}-\hat{\boldsymbol{\delta}}_{R} \|_2^2
\end{align}
where $\boldsymbol{\delta}_{R}$ represents the vector column of the values obtained from (\ref{eq:rotation_corr}) and $\boldsymbol{\hat{\delta}}_{R}$ is given by (\ref{eq:rotation_linear_K_matrix}). The solution is \begin{align}\label{eq: optimize_vSol}
\mathbf{V}=(\mathbf{A}^T\mathbf{A})^{-1}\mathbf{A}^T{\hat{\boldsymbol{\delta}}}_{R}.
\end{align}

After the relation between the considered property (rotation) and latent code is established, we can now use it to calculate a satisfying value of the latent code $c_1$ to produce a SAR image, $\mathbf{I}_d$, for any desired rotation angle $\delta_{Rd}$,
\begin{align}
	c_1 = \frac{\delta_{Rd}-v_0}{v_1},
\end{align}
as $\mathbf{I}_d=\mathbf{G}(\mathbf{z},c_1)$.

Linear model is very simple, however, as will be seen from the experiments, it can be used as a rough model only. Namely, the true relation between rotation and latent code is nonlinear, being governed by nonlinearities in the InfoGAN.     

(b) \textit{Nonlinear model}:  From the experiments, we concluded that a general form of a function (following the sigmoid function at the output of the neural network) is quite an appropriate model for the relation between the physical properties of the SAR image and the latent codes. The sigmoid follows the $\tanh$ function. A nonlinear model of, for example, rotation, with one latent code $\mathbf{c}_1$ could be written as:     
\begin{align}\label{eq:rotationtanh_1c}
	\hat{\delta}_R(k) = v_3\tanh(v_{1}c_1(k)+v_2) + v_{0}, \ \ \ k=1,2,\dots,K.
\end{align}
The solution to the minimization problem (\ref{eq: optimize_v}) cannot be obtained in analytic form, for this case. However, the tools for numerical solution to this  problem are well developed in all programming environments. Therefore, we may say that the values of $\mathbf{V}=	[v_0, \ v_1, \ v_2, \ v_3]^T$ can be obtained from a set of $k$ nonlinear equations in (\ref{eq:rotationtanh_1c}). After the model coefficients, $\mathbf{V}$, are found, we can again easily find a latent code $c_1$ to generate a SAR image, $\mathbf{I}_d$, with a desired parameter 
 $\delta_{Rd}$, as
\begin{align}\label{inversetanh}
	c_1 = \frac{1}{v_1}\tanh^{-1}\Big(\frac{\delta_{Rd}}{v_3}-v_0\Big)-v_2.
\end{align}
as $\mathbf{I}_d=\mathbf{G}(\mathbf{z},c_1)$.

\noindent\textbf{ One property - Two latent codes:} In SAR images, after the basic property change, we can expect other changes to occur as well (like changes in the reflectivity and visibility of scatters). This means that even with one geometric property change, we may still use more than one latent code. Now we extend the analysis to two latent codes $c_1$ and $c_2$. The linear model for two latent code space can be expressed as
\begin{align}\label{eq:rotationlinear_2c}
	\hat{\delta}_R(k_1,k_2) = v_{2}c_2(k_2) +v_{1}c_1(k_1) + v_{0}, \ \  k_1,k_2=1,2,\dots,K. \nonumber
\end{align}
If we form a stacked column vector $\hat{\boldsymbol{\delta}}_{R}$ with $K^2$ elements $\hat{\delta}_R(k_1,k_2)$, $K^2\times 3$ matrix $\mathbf{A}$ with rows $[c_2(k_2), c_1(k_1), 1]$,  and the column vector of unknown coefficients $\mathbf{V}=	[v_2, \ v_1, \ v_0]^T$, then the solution is again obtained in the form $\mathbf{V}=(\mathbf{A}^T\mathbf{A})^{-1}\mathbf{A}^T{\hat{\boldsymbol{\delta}}}_{R}$.

In this case, the latent code values for a given property, for example rotation $\delta_{Rd}$, is not unique since all combinations of the latent codes along the line  
\begin{equation}\label{eq:inv1p2c}
	 v_{2}c_2 +v_{1}c_1 =v_{0}-\delta_{Rd}
\end{equation}
in the $c_1$-$c_2$ plane which will produce the same desired rotation $\delta_{Rd} $.
The desired rotation can be obtained by fixing one latent code, $c_1$ or $c_2$, and calculating the other latent code value.

For two latent codes, the nonlinear model is of the form
\begin{gather}\label{eq:rotationtanh_2c}
	\hat{\delta}_R(k_1,k_2) = {}  v_4\tanh(v_{1}c_1(k_1)+  v_2c_2(k_2) + v_3) + v_{0}, \\ 
	 k_1,k_2=1,2,\dots,K  \nonumber
\end{gather}
The optimization of parameters $v_4$, $v_3$, $v_2$, $v_1$, and $v_0$, is done using  common nonlinear fitting tools.
The line for a desired $\delta_{Rd}$ is obtained in the form
\begin{align}\label{eq:rotationtanh_2cc}
v_{1}c_1+  v_2c_2= \tanh^{-1}\Big(\frac{\delta_{Rd}-v_0}{v_4}\Big).
\end{align}
Again, a desired $\delta_{Rd}$ can be achieved with all pairs of $(c_1,c_2)$ on the previous line.

In the nonlinear model, we further introduce a quadratic term in the argument of the $\tanh$ function as
\begin{gather}\label{eq:rotationtanh_2c_quadratic}
		\delta_R(k_1,k_2) = v_7\tanh(P_R(c_1(k),c_2(k_2)) + v_{0}, \\  
		k_1,k_2=1,2,\dots,K.  \nonumber
\end{gather}  
where $P_R(c_1(k),c_2(k_2))=  v_1c^2_1(k_1)+v_2c^2_2(k_2)+v_3c_1(k_1)c_2(k_2)+v_4c_1(k),+v_5c_2(k)+v_6$, $k_1,k_2=1,2,\dots,K$. For a desired $\delta_{Rd}$, ($c_1$, $c_2$) should be satisfied the following relation
\begin{align}\label{eq:inv1p2cquadratic}
	P_R(c_1, c_2) = \tanh^{-1}\left(\frac{\delta_{Rd-v_0}}{v_7}\right)
\end{align}
meaning all combinations of the latent codes are along a quadratic form line. Namely, (\ref{eq:inv1p2cquadratic}) is a  general quadratic equation, producing conic sections (circles, ellipses, parabolas, and hyperbolas) in the $c_1$-$c_2$ plane, depending on the specific parameter $v_0,v_1,v_2,\dots,v_7$ values.

\noindent\textbf{ Two properties - Two latent codes:}
For a simultaneous change of two properties, we will use two codes and a nonlinear model. In the nonlinear model, we will use a linear argument form of the $\tanh$ function and a quadratic argument of this function. In the case of the linear argument, we will use the model 
\begin{align}\label{eq:rotationtanh_2c2c}
	\delta_R(k_1,k_2) & = v_4\tanh(v_{1}c_1(k_1)+  v_2c_2(k_2) + v_3) + v_{0}, \\  
		\delta_S(k_1,k_2) & = v_9\tanh(v_{6}c_1(k_1)+  v_7c_2(k_2) + v_8) + v_{5},  \nonumber
\end{align}  
The quadratic argument model is of the form
\begin{align}\label{eq:rotationtanh_2c2cQ}
	\delta_R(k_1,k_2) & = v_7\tanh(P_R(c_1(k),c_2(k_2)) + v_{0}, \\  
	\delta_S(k_1,k_2) & = v_{15}\tanh(P_S(c_1(k),c_2(k_2)) + v_{8}, \\
& \qquad	k_1,k_2=1,2,\dots,K, \nonumber
\end{align}  
where the polynomial arguments for the two properties are defined by
\begin{align}\label{eq:rotationtanh_2c2cQP}
	P_R(c_1(k_1),c_2(k_2))={} & v_1c^2_1(k_1)+v_2c^2_2(k_2)+v_3c_1(k_1)c_2(k_2) \\ + v_4c_1(k_1)+v_5(c(k2))+v_6, \\  	
	P_S(c_1(k_1),c_2(k_2))={} & v_9c^2_1(k_1)+v_{10}c^2_2(k_2)+v_{11}c_1(k_1)c_2(k_2) \\ + v_{12}c_1(k_1)+v_{13}(c(k2))+v_{14},   
\end{align} 
for 	$k_1,k_2=1,2,\dots,K$. 
These two systems are independently solved for the corresponding sets of coefficients in the model. 

In this case, the desired SAR image is generated at the intersection of the lines producing desired rotation, $\delta_{Rd}$, and scaling, $\delta_{Sd}$, since for each of them we get the corresponding lines as in 
(\ref{eq:rotationtanh_2cc}) and (\ref{eq:inv1p2cquadratic}).

All the previous setups will be illustrated and explained in more details in the next section dealing with experimental results.

\section{Experiments}\label{sec:Experiment}

\begin{figure}[tbp]
	\centering
	\includegraphics[]{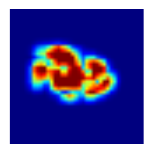}
	\includegraphics[]{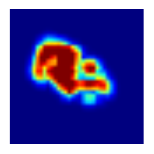}
	\includegraphics[]{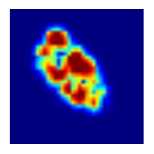}
	\includegraphics[]{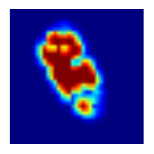}
	\includegraphics[]{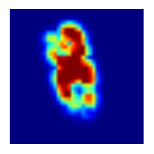}
	
	\includegraphics[]{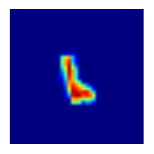}
	\includegraphics[]{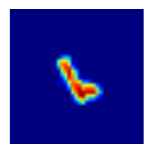}
	\includegraphics[]{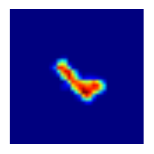}
	\includegraphics[]{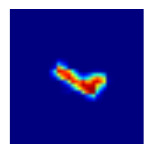}
	\includegraphics[]{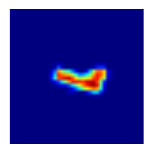}
	
	\includegraphics[]{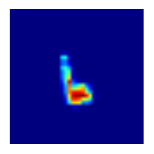}
	\includegraphics[]{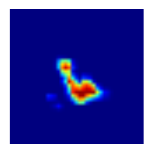}
	\includegraphics[]{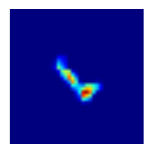}
	\includegraphics[]{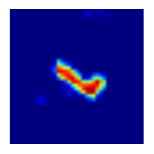}
	\includegraphics[]{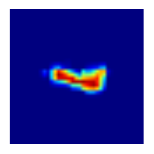}
	
	\includegraphics[]{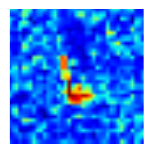}
	\includegraphics[]{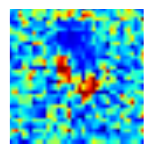}
	\includegraphics[]{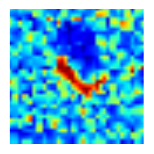}
	\includegraphics[]{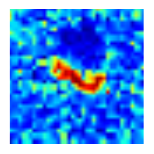}
	\includegraphics[]{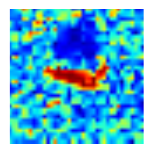}
	\caption{Illustration of SAR image samples from four data sets considered in the experimental setup:  Simulated SAR images with different viewing angles (top row). A radar image from the MSTAR dataset, with suppressed background, rotated for various angles (second row).  SAR images from MSTAR dataset corresponding to different viewing angles of the same target, with suppressed background (third row).  SAR images from MSTAR dataset corresponding to different viewing angles with a background (bottom row).}
	\label{dataset}
\end{figure}

In our experiments, four kinds of datasets are utilized: 
\begin{itemize}
	\item simulated images
	\item real object from a SAR image with simulated properties
	\item real SAR images with suppressed background
	\item real SAR images with background
\end{itemize}
as shown in Fig.~\ref{dataset}.

Now we will introduce the architecture of InfoGAN in our experiments in detail before various experiments. 


		

\medskip
		
\noindent\textbf{InfoGAN Architecture}: The generator $G$ contains one fully-connected layer and four transposed convolutional layers. 
The input $\mathbf{z}$ to the generator is a one-dimensional vector concatenating pure noise, and latent codes in the length of $N_z$ ($N_z=N_N+N_C$), where $N_N$, $N_C$ denote the length of noise and latent codes. Unless specified, $N_z=62$ in this paper. $N_C$ equals the number of classes and latent codes.
The discriminator $D$ contains four convolutional layers and one fully-connected layer. The classifier $Q$ contains four convolutional layers and two fully-connected layers.   $D$ and $Q$ share the parameters for all convolutional layers. 
In our experiments, there are two latent codes at most, thus two single neurons are set in the output layer of $Q$. Table. \ref{tab:res1} and Table. \ref{tab:res2} show the details of $G$,  $D$, and  $Q$, respectively. 
To avoid modifying InfoGAN's architecture, we assign a $0$ weight to the loss function of the second one of two latent codes when only one latent code is required.

\begin{table}[tbp]
\centering
	\caption{The architecture of the generator, $G$ \label{tab:res1}}
	\begin{tabular}{ccccc}
		\toprule  
		Layer&Input shape&Output shape&Activation\\ 
		\midrule  
		Fully-connected &$N_z$	&  $6272$ & \\
		Reshape &$6272$	&  $7\times7\times128$  & \\
		BatchNormalize &$7\times7\times128$ &  $7\times7\times128$ & Sigmoid \\
		TransposedConv2D &$7\times7\times128$	& $14\times14\times128$ & \\	
		BatchNormalize &$14\times14\times128$ 	& $14\times14\times128$  &Sigmoid \\	
		TransposedConv2D &$14\times14\times128$ 	&$28\times28\times64$   & \\	
		BatchNormalize &$28\times28\times64$ 	& $28\times28\times64$  & Sigmoid \\	
		TransposedConv2D &$28\times28\times64$ 	& $28\times28\times32$  & \\	
		BatchNormalize &$28\times28\times32$ 	& $28\times28\times32$  & Sigmoid \\	
		TransposedConv2D &$28\times28\times32$ 	& $28\times28\times1$  & Sigmoid \\	
		\bottomrule  
	\end{tabular}
\end{table}

	\begin{table}[tbp]
			\centering
			\caption{The architecture of the discriminator $D$ and the classifier, $Q$  \label{tab:res2}}
			\begin{tabular}{cccccc}
					\toprule  
					Layer&Input shape&Output shape&Activation\\ 
					\midrule  
					Conv2D & $28\times28\times1$	&  $14\times14\times32$ & Leaky ReLU\\
					Conv2D & $14\times14\times32$	&  $7\times7\times64$ &  Leaky ReLU\\
					Conv2D & $7\times7\times64$	&  $4\times4\times128$ &  Leaky ReLU\\
					Conv2D & $4\times4\times128$	&  $4\times4\times256$ &  Leaky ReLU\\
					Flatten & $4\times4\times256$	&  $4096$ & \\
					\midrule  
					$D$: Fully-connected &$4096$	&  $1$ &  Sigmoid\\
					\midrule  
					$Q$: Fully-connected &$4096$	&  $128$ &  \\
					~~~~~~Fully-connected &$128$	&  $N_C$ & Sigmoid\\
					
					\bottomrule  
				\end{tabular}
		\end{table}
	
In the following experiments, the simulated images are of size $28 \times 28$ pixels, while the real data images are downsampled to this size. The learning process for InfoGAN lasted about $10$ minutes with $10000$ iterations on a laptop computer with a CPU of 3.2GHz, RAM of 32 GB, and GPU NVIDIA Geforce RTX 3070. Larger images can be processed in the same waywith some increase in the computation time.

\subsection{Simulated SAR Images}

 The SAR images of a ship are simulated in this experiment. The radar operating frequency $f_0=157GHz$, $T_r=93.75\mu s$, with $28$ pulses and $28$ range cells inside a pulse.  The target is illuminated from different angles (or the target is rotated) with an angle from $10$ to $70$ degrees with respect to the line of flight. 
For the first experiment, only the rotation is considered since it is the most complex property for simulated SAR images as discussed in Section II. 

\begin{figure}[tbp]
	\centering
	\includegraphics[]{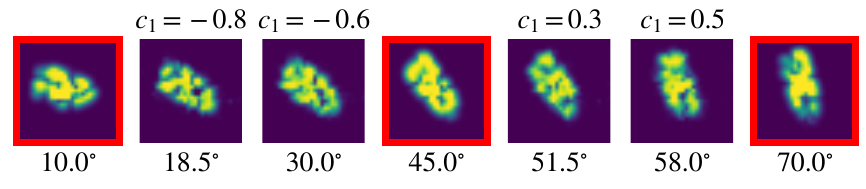}
	\caption{Real and synthesized SAR images for various rotation angles. The first, fourth, and seventh images (marked by red square) are SAR images used for the training of the InfoGAN, while the second, third, fifth, and sixth images are the SAR images synthesized by the InfoGAN with the latent code values $c_1=-0.8, -0.6, 0.3, 0.5$, respectively.}
	\label{Simnotintraining}
\end{figure}

The InfoGAN described above (Tables \ref{tab:res1} and \ref{tab:res2}) is trained with only one latent code, $c_1$, activated.  For the beginning, only $13$ training images ($5^{\circ}$ step) are used to train the InfoGAN. After the  InfoGan is trained, we have tested various values of $c_1$ and generated new SAR images. The resulting images covered almost the whole rotation angle range. This means that some rotation angles not appearing in training can be synthesized by manipulating the latent code $c_1$, values, with examples as shown in Fig.~\ref{Simnotintraining}.

\begin{figure}[tbp]
	\centering
	{\includegraphics{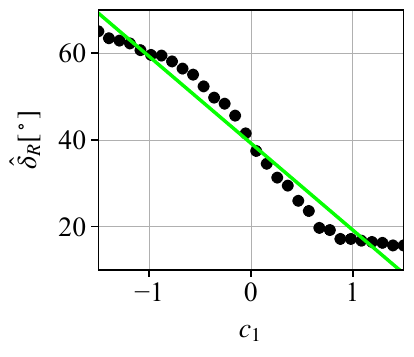}}
	\includegraphics[]{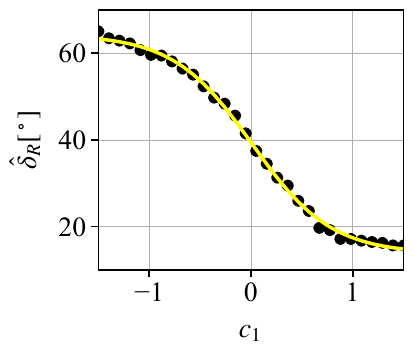}
	
	{\includegraphics{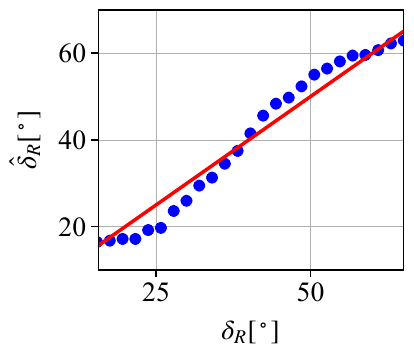}}
	{\includegraphics{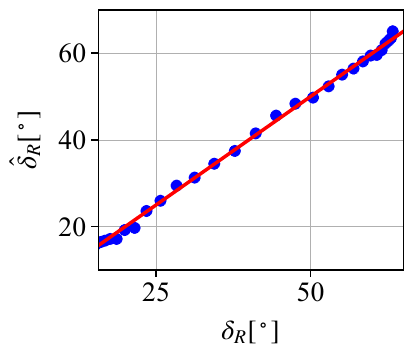}}
	
	{\includegraphics{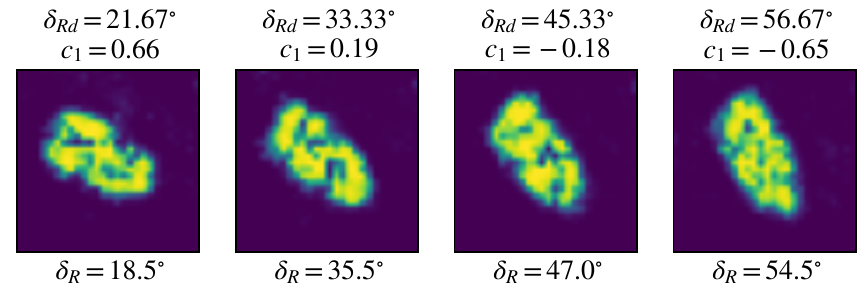}}
	\caption{The results for the estimated and modeled rotation angle for the SAR images synthesized by the InfoGAN trained with simulated SAR images.  The rotation angles in SAR image as a function of the latent code, $c_1$, measured by cross-correlation (black dots) and the estimated values with a linear model (green line) (top-left).  The rotation angles in SAR image as a function of the latent code, $c_1$, measured by cross-correlation (black dots) and the estimated values with a nonlinear model (yellow line) (top-right).  Comparison of the measured angle values by cross-correlation with the ones obtained using the linear model (blue dots), where the red line denotes the ideal case that $\hat{\delta}_{R}(k)=\delta_{R}(k)$ for all $k$ (middle-left).  Comparison of the estimated angle values with the ones obtained using the nonlinear model (blue dots) (middle-right).  The synthesized SAR images using $c_1$ calculated by (\ref{inversetanh}) for four desired rotation angles, $\delta_{Rd}=21.67^{\circ}, \ 33.33^{\circ}, \ 45.33^{\circ},  \ 56.67^{\circ}$ (bottom row). The estimated rotations of the synthesized SAR images, $\delta_R(k)$ are calculated using (\ref{eq:rotation_corr}). They are close to the desired ones. }
	\label{sim_acc}
\end{figure} 	


For a detailed analysis of the relation between the rotation angle, $\delta_R$,  and the latent code, $c_1$, the number of training images is increased to $121$ within the same range from $10$ to $70$ degrees with respect to the line of flight.

After the InfoGAN is trained, we have generated a set of images corresponding to the various values of the latent code, $c_1(1), \cdots, c_1(K), ~K=30$, uniformly sampled from the interval $[-1.5,1.5]$.  After the SAR images are synthesized using these latent code values, the rotation angles, $\delta_R(k)$, $k=1,2,\dots,K$, are measured for the obtained SAR images with each latent code, using (\ref{eq:rotation_corr}), and the parameters $\mathbf{V}$ of a linear and nonlinear model are calculated by equation (\ref{eq: optimize_vSol}) or solving the system (\ref{eq:rotationtanh_1c}), respectively. The liner model solution is show in the Fig.~\ref{sim_acc}(top-left) with a green line, while the measured angles $\delta_R(k)$ are given by dots. This panel shows that the rotation angle changes in approximately linear way with respect to $c_1$. A direct comparison of  the measured angle,  $\delta_R(k)$, and the estimated angle by a linear model,  $\hat{\delta}_R(k)$, is shown in Fig.~\ref{sim_acc}(bottom-left). The procedure is repeated with the nonlinear model (\ref{eq:rotationtanh_1c}) and the corresponding results are shown  Fig.~\ref{sim_acc}(top-right) and Fig.~\ref{sim_acc}(bottom-right).
It is clear that nonlinear model  performs better than the linear model, which will be even more evident in the next experiments. 

Finally, the model is tested with four desired rotation angles, $\delta_{Rd}=21.67^{\circ}$, $ 33.33^{\circ}$, $45.33^{\circ}$, $56.67^{\circ}$. The latent code values, $c_1$, for these rotations are calculated using  (\ref{inversetanh}). Then the InfoGAN produced the synthesized SAR images, shown in Fig.~\ref{sim_acc} (bottom row). The estimated rotations $\delta_R(k)$ are obtained from (\ref{eq:rotation_corr}). They are within a few degrees margin with respect to the desired ones.

\subsection{Real object from a SAR Image with Simulated Properties}	

After the simulated SAR examples, before a real data example, as an intermediate step, we shall consider a SAR image from the real data set MSTAR \colcit{\cite{MSTAR}} (a popular public SAR image dataset which will be elaborated in next subsection), but to fully control the transformations, we will produce new images by rotating, scaling, and shifting the assumed real SAR image. Unless otherwise specified, the background in each SAR image has been removed before all experiments by using Self-Matching CAM \colcit{\cite{feng2021self}}. Recall that geometrical transformations will be, in general, referred to the properties. As in Section \ref{sec:Methodology}, we set three cases for the considered images and the InfoGAN: (1) One property - One latent code; (2) One property - Two latent codes; (3) Two properties - Two latent codes.

\subsubsection{One property - One latent code}

\begin{figure}[tbp]
	\centering
{\includegraphics{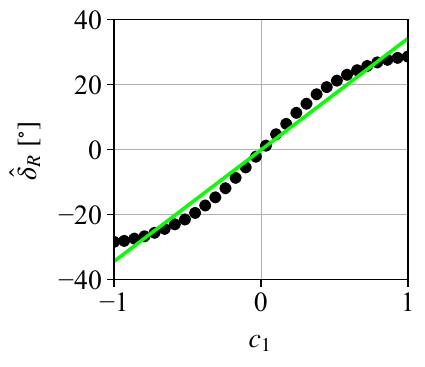}}%
{\includegraphics{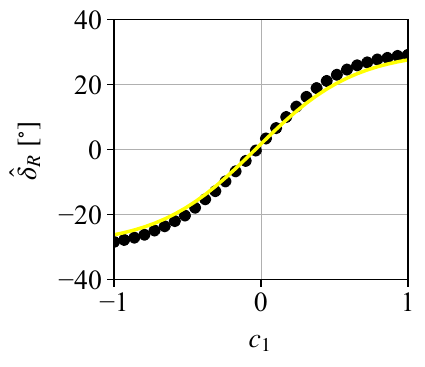}}
	
	{\includegraphics{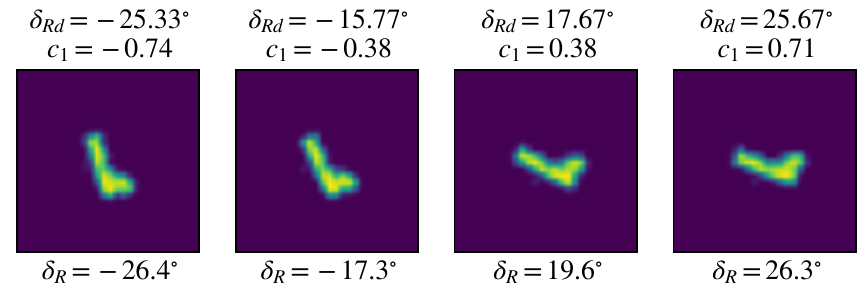}}
\smallskip
	
{\includegraphics{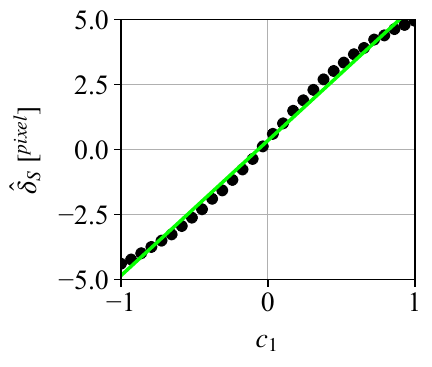}}%
{\includegraphics{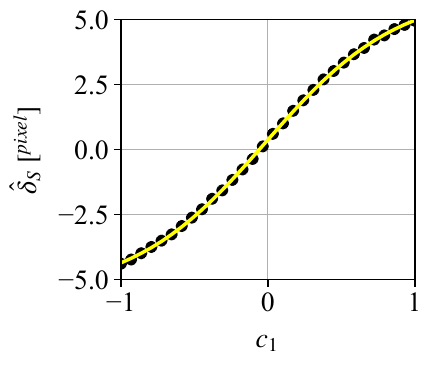}}
	
	{\includegraphics{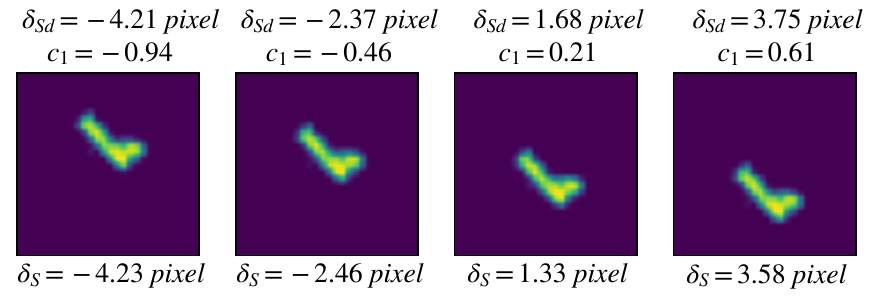}}
\smallskip
	
{\includegraphics{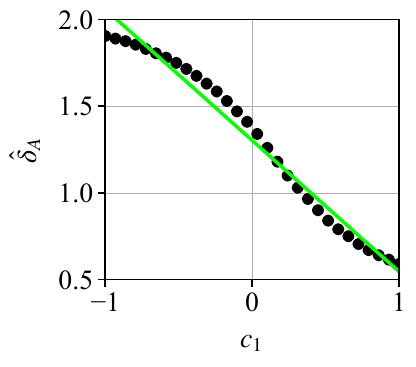}}%
{\includegraphics{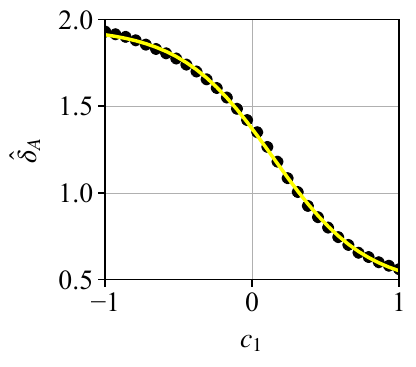}}
	
	{\includegraphics{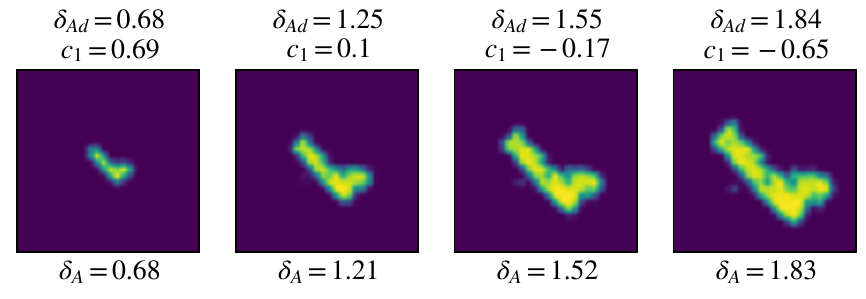}}
	\caption{The results for the measured and  modeled rotation (top), translation (middle), and scaling (bottom)  for the SAR images synthesized by the InfoGAN trained with the second dataset
		For each case we show the relation between $c_1$ and the considered property (dots), approximations using linear (green line in left subplots) and nonlinear model (yellow line in right subplots),
 and synthesized SAR images using $c_1$ calculated by (\ref{inversetanh}) for four desired $\delta_{Rd}$, $\delta_{Sd}$, and $\delta_{Ad}$. The estimated properties of the synthesized SAR images,  $\delta_{R}$, $\delta_{S}$, and $\delta_{A}$ are measured by (\ref{eq:rotation_corr}). They are close to desired ones.}
	\label{1p1caccF}
\end{figure}

All three properties are considered separately: for rotation, a real SAR image is analytically rotated from $-30$ to $30$ degrees to obtain $601$ images; for translation, the target in real image is translated from $-6$ to $6$ pixels from the original position to obtain $151$ images; for scaling,  the target in real image is scaled from $0.5$ to $2$ times of the original size to obtain $301$ images. After the InfoGAN is trained independently with three datasets, respectively (in three separate experiments), we have synthesized the new images corresponding to the various values of the latent code, $c_1(1), \cdots, c_1(K), ~K=30$, uniformly sampled from the interval $[-1.0,1.0]$ for each property.  Then the properties, $\delta_{R}$, $\delta_{S}$, $\delta_{A}$ can be measured by (\ref{eq:rotation_corr}) and the estimated properties, $\hat{\delta}_{R}$, $\hat{\delta}_{S}$, $\hat{\delta}_{A}$, can calculated using (\ref{eq:rotationLinear11}) and (\ref{eq:rotationtanh_1c}).  The comparison of the measured properties and estimated properties shows that the nonlinear estimator  performs better than linear estimator in all cases, especially for rotation (top-right) and scaling (bottom-right) in Fig.~\ref{1p1caccF}.  For each case, we have synthesized SAR images  for four desired $\delta_{Rd}$, $\delta_{Sd}$, and $\delta_{Ad}$, respectively, using $c_1$ calculated by (\ref{inversetanh}). The estimated properties of the synthesized SAR images,  $\delta_{R}$, $\delta_{S}$, and $\delta_{A}$ are measured by (\ref{eq:rotation_corr}). We can see that the agreement is good in all considered cases.

\subsubsection{One property - Two latent codes}

\begin{figure}[tpb]
	\centering
	{\includegraphics{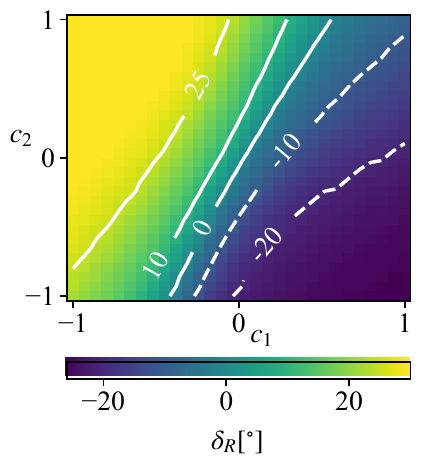}}
	{\includegraphics{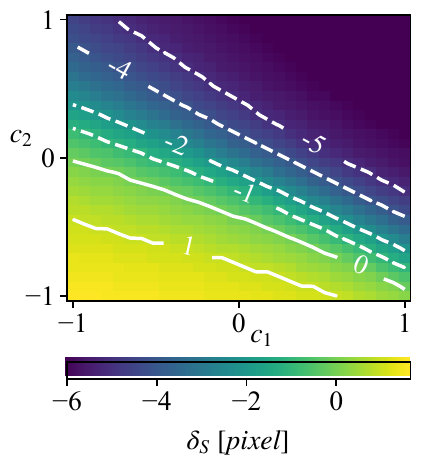}}
	
	{\includegraphics{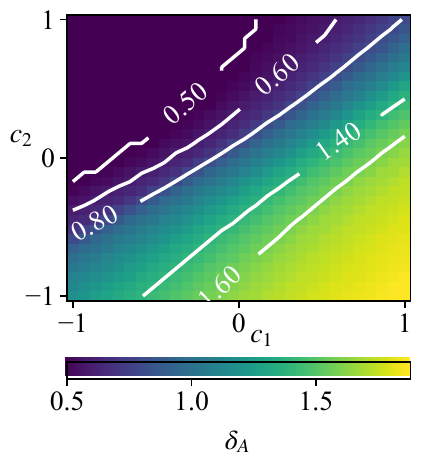}}
	{\includegraphics{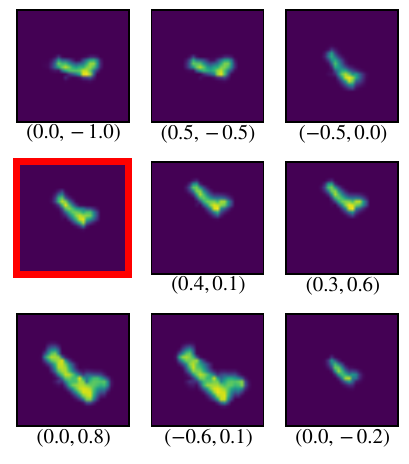}}
	\caption{{
			The relation between each property and two latent codes.  The relation between rotation angle $\delta_R$ and $c_1$, $c_2$ (top-left).  The relation between translation pixels $\delta_S$ and $c_1$, $c_2$ (top-right).  The relation between scaling $\delta_A$ and $c_1$, $c_2$ (bottom-left).  The synthesized SAR images corresponding to ($c_1$, $c_2$) labeled below each image except for a original image (marked by red square) (bottom-right). In this panel (bottom-right), the first two images the top row exhibit the same rotation angle $\delta_{R}$ with different $c_1$ and $c_2$, i.e., $c_1=0.0$, $c_2=-1.0$ and $c_1=0.5$, $c_2=-0.5$ both resulting in $-20^{\circ}$ rotation. The third one in the top row shows $\delta_R=25^{\circ}$ with $c_1=-0.5$ and $c_2=0.0$. These figures furthers demonstrate the solution to (\ref{eq:inv1p2cquadratic}) is not unique, thus it is possible to retain or change property by manipulating $c_1$ and $c_2$. This conclusion is also applicable to translation $\delta_{S}$ and scaling $\delta_{A}$, as shown in  the second and the third rows in (bottom-right).  
	}} 
	\label{1p1cacc}
\end{figure} 

\begin{figure}[tbp]
	\centering
	{\includegraphics{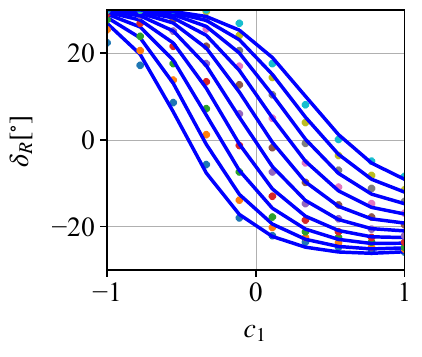}}
	{\includegraphics{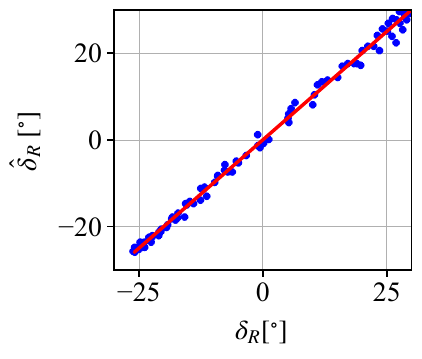}}
	\hspace*{1.5mm}{\includegraphics{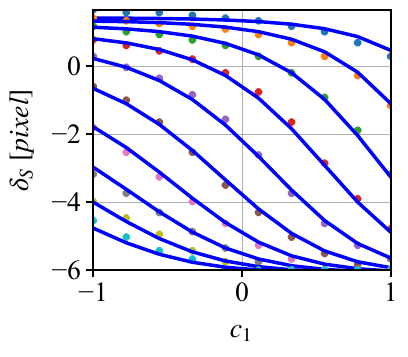}}
	{\includegraphics{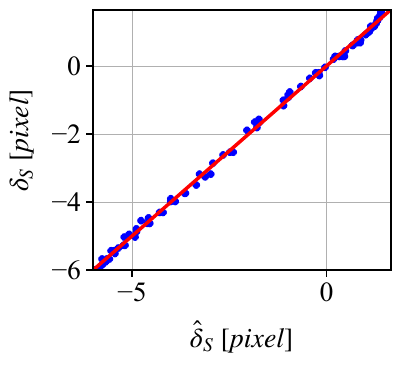}}	
	{\includegraphics{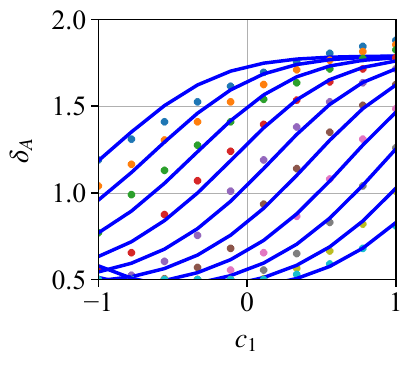}}
	{\includegraphics{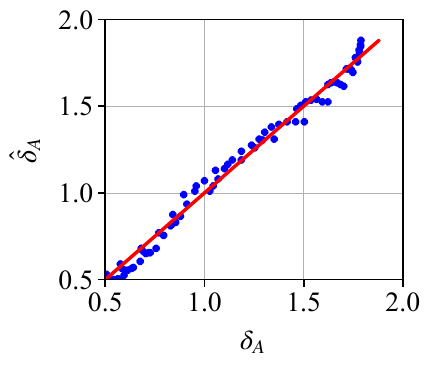}	
	}
	\caption{The comparison of three estimated properties $\hat{\delta}_R$, $\hat{\delta}_S$, and $\hat{\delta}_A$, using (\ref{eq:rotationtanh_2c_quadratic}), and the measured ones, $\delta_R$, $\delta_S$, and $\delta_A$, using (\ref{eq:rotation_corr}).  The relation with $\delta_R$ (dots) and two latent codes, $c_1$ and $c_2$ (different colors denotes different values of $c_2$) (top-left). The $\hat{\delta}_{R}$ is shown with blue lines. They are close to the $\delta_R$.  The comparison of $\delta_R$ and $\hat{\delta}_R$ (top-right).  The results of $\delta_S$ and $\hat{\delta}_S$ are shown in (middle-right) and (middle-left), respectively. The results of $\delta_A$ and $\hat{\delta}_A$ are shown in (bottom-right) and (bottom-left), respectively.  }
	\label{1p2cacc}
\end{figure} 

Now we introduce two latent codes $c_1$ and $c_2$ to train the InfoGAN with input images exhibiting one property variations in order to check one property will remain within one latent code or will propagate to the other latent code as well. The setup of training data is completely the same as in the previous experiment. Take rotation as instance, we have generated $900$ images with $\delta_{R}(k_1,k_2)$, $k_1,k_2=1,2,\dots,30$, from the InfoGAN trained with both $c_1$ and $c_2$ activated. Fig.~\ref{1p1cacc} reveals that the value of a specific property is spread over the available latent codes and therefore is determined  by multiple pairs of $c_1$ and $c_2$, because the solution to (\ref{eq:inv1p2cquadratic}) is not unique, as discussed in Section \ref{sec:Methodology}. 

To show this relation vividly, we generated several SAR images by using some selected values of $c_1$ and $c_2$, as shown in Fig.~\ref{1p1cacc} (bottom-right). In this panel, consisting $3\times 3$ images, the first and the second image in the top row are with different $c_1$ and $c_2$ but both resulting in the same $\delta_R=-20^{\circ}$. In comparison, the third one in the top row shows $\delta_R=25^{\circ}$ with $c_1=-0.5$ and $c_2=0.0$. This comparison further demonstrates the solution to (\ref{eq:inv1p2c}) is not unique. This conclusion is also applicable to $\delta_{S}$ and $\delta_{A}$ as shown in the second and the third row, thus it is feasible to retain or change any property by manipulating $c_1$ and $c_2$.
Finally, the properties measured by (\ref{eq:rotation_corr}) and the estimated properties using (\ref{eq:rotationtanh_2c_quadratic}) are compared in Fig.~\ref{1p2cacc} to validate the performance of the estimator (only nonlinear model is considered because the relation between one property and two latent codes is obviously much more complex than linear model). The results show that $\hat{\delta}_R$,  $\hat{\delta}_S$, and  $\hat{\delta}_A$, calculated by (\ref{eq:rotationtanh_2c_quadratic}) basically match the $\delta_R$, $\delta_S$, and $\delta_A$, respectively, even though the accuracy is slightly lower than in Fig.~\ref{1p1cacc}.

\subsubsection{Two properties - Two latent codes}
\begin{figure}[tbp]
	\centering
	{\includegraphics{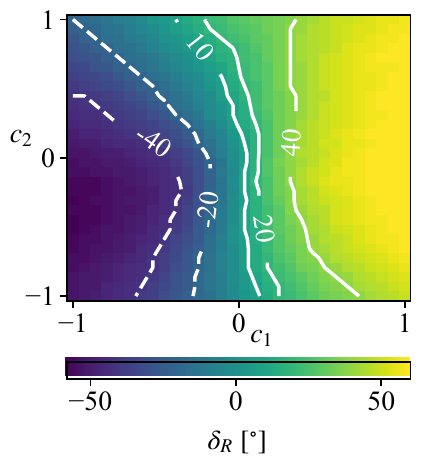}}
	{\includegraphics{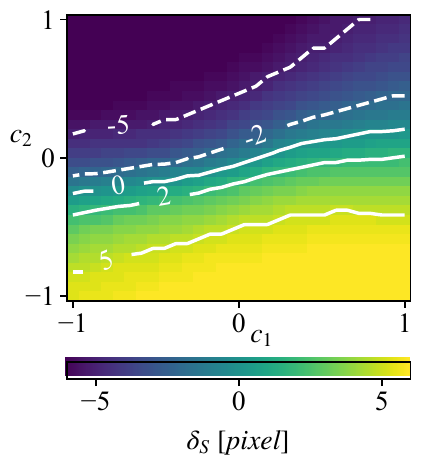}}
	
	{\includegraphics{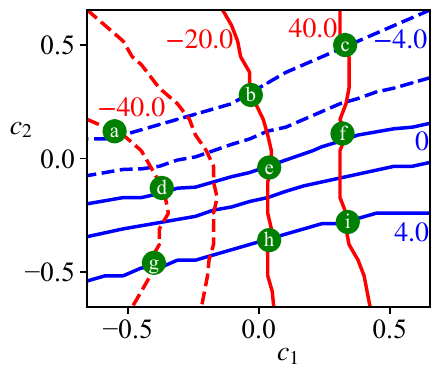}}
	{\includegraphics{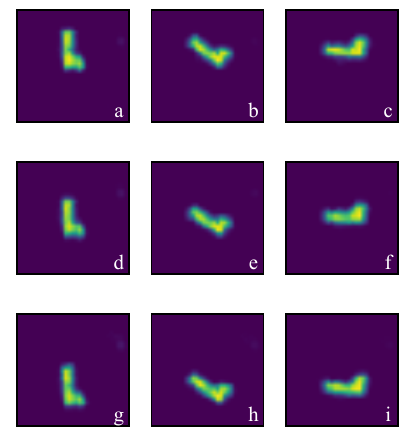}}
	\caption{The relation between rotation-translation and two latent codes. The relation between rotation angle $\delta_R$ and $c_1$, $c_2$, (top-left). The relation between translation $\delta_S$ and $c_1$, $c_2$, (top-right). The overlapped curves of the above two contours as well as some selected intersections (green dots) (bottom-left). The synthesized SAR images with ($c_1$, $c_2$) corresponding to the coordinates of the green dots in the former contour (bottom-right). } 
	\label{fig:2p2c_rotpos}
\end{figure}

\begin{figure}[tbp]
	\centering
	{\includegraphics{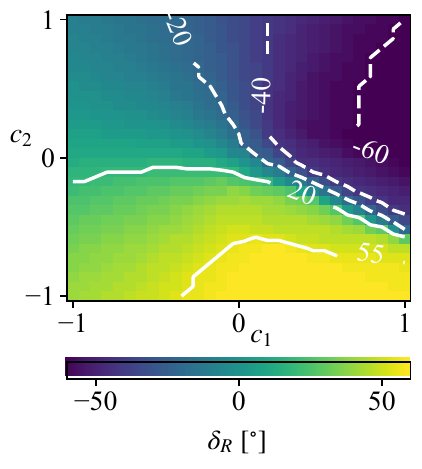}}
	{\includegraphics{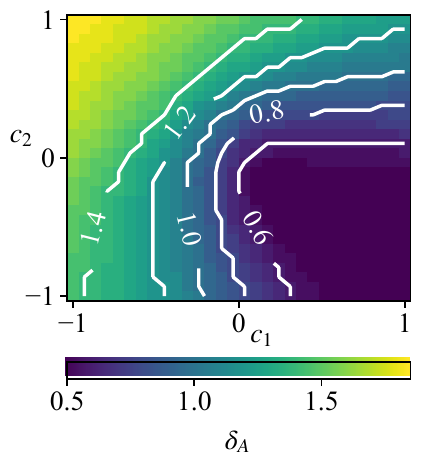}}
	
	{\includegraphics{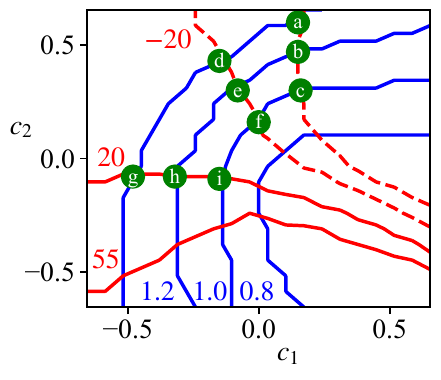}}
	{\includegraphics{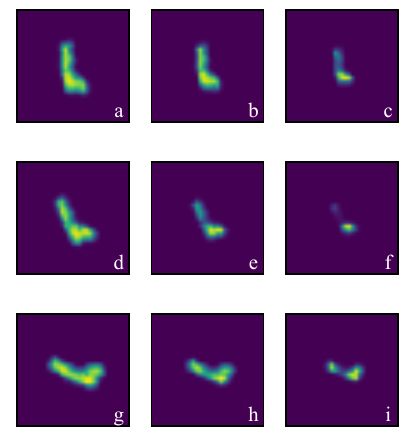}}
	\caption{The relation between rotation-scaling and two latent codes. The relation between rotation angle $\delta_R$ and $c_1$, $c_2$  (top-left).  The relation between scaling $\delta_A$ and $c_1$, $c_2$ (top-right).  The overlapped curves of the above two contours as well as some selected intersections (green dots) (bottom-left).  The synthesized SAR images with ($c_1$, $c_2$) corresponding to the coordinates of the green dots in the former contour (bottom-right). } 
	\label{fig:2p2c_rotsc}
\end{figure}

\begin{figure}[tbp]
	\centering
	{\includegraphics{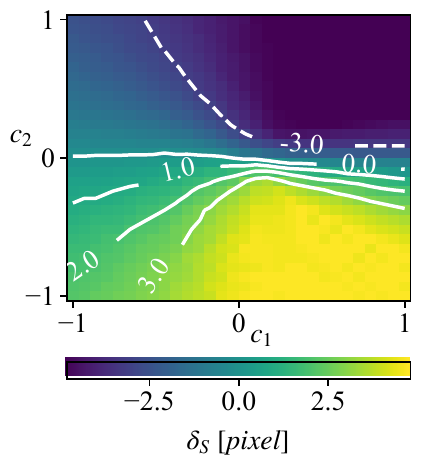}}
	{\includegraphics{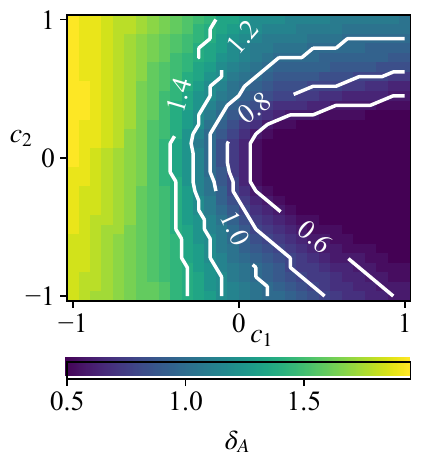}}
	
	{\includegraphics{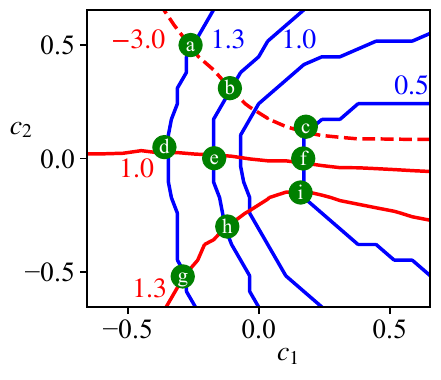}}
	{\includegraphics{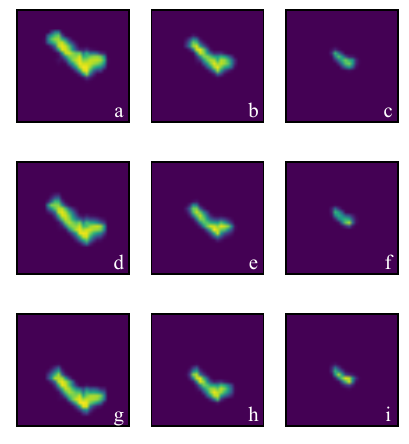}}
	\caption{The relation between translation-scaling and two latent codes. The relation between translation angle $\delta_S$ and $c_1$, $c_2$  (top-left). The relation between scaling $\delta_A$ and $c_1$, $c_2$ (top-right). The overlapped curves of the above two contours as well as some selected intersections (green dots)  (bottom-left).  The synthesized SAR images with ($c_1$, $c_2$) corresponding to the coordinates of the green dots in the former contour (bottom-right). } 
	\label{fig:2p2c_scpos}
\end{figure}

In this experiment, we consider two entangled properties emerging in each training SAR image simultaneously. Firstly, we generate three combinations of training data: rotation-translation, rotation-scaling, and translation-scaling. For rotation-translation, there are $3721$ training images with $61$ rotation angles uniformly dividing $[-60^{\circ}, 60^{\circ}]$ and $61$ translation pixels uniformly dividing $[-6, 6]$ pixel.  For rotation-scaling, there are $1891$ training images with $31$ scaling uniformly dividing $[0.5, 2]$ and $61$ rotation angles uniformly dividing $[-60^{\circ}, 60^{\circ}]$. For translation-scaling, there are $3751$ training images with $121$ translation pixels uniformly dividing $[-6, 6]$ pixel and $31$ scaling uniformly dividing $[0.5, 2]$. 
We have generated $900$ images for each property using different combinations of $c_1$ and $c_2$ and show their relation in 
Fig.~\ref{fig:2p2c_rotpos}, Fig.~\ref{fig:2p2c_rotsc}, and Fig.~\ref{fig:2p2c_scpos}.
Next, we conduct an experiment to visualize how to edit the entangled properties by manipulating $c_1$ and $c_2$. 
In each case, we select $9$ combinations of $c_1$ and $c_2$ in intersections of two contour lines (green dots in (bottom-left) in Figs.~\ref{fig:2p2c_rotpos}, \ref{fig:2p2c_rotsc}, and \ref{fig:2p2c_scpos}).  The synthesized SAR images by using these ($c_1$, $c_2$) in (bottom-right) show that if $c_1$ and $c_2$ are along one curve, only the property corresponding to this curve will be changed while the other property remains still.
Furthermore, given two desired properties, for example, $\delta_{Rd}$ and $\delta_{Sd}$, the satisfying combination of $c_1$ and $c_2$ is unique in a certain range (the green dots).  Thus, it is feasible to precisely edit either single property or two properties simultaneously by manipulating $c_1$ and  $c_2$ as we have expected.

\subsection{Real SAR Images with Suppressed Background}

\begin{figure}[tpb]	
	\centering
	\includegraphics{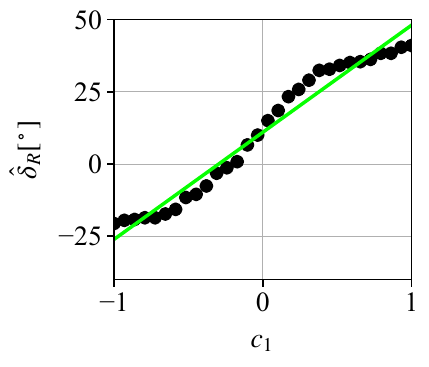}
	{\includegraphics{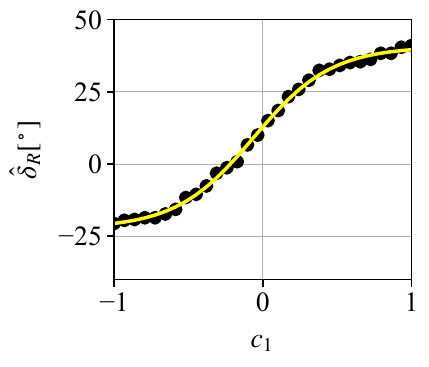}}
	{\includegraphics{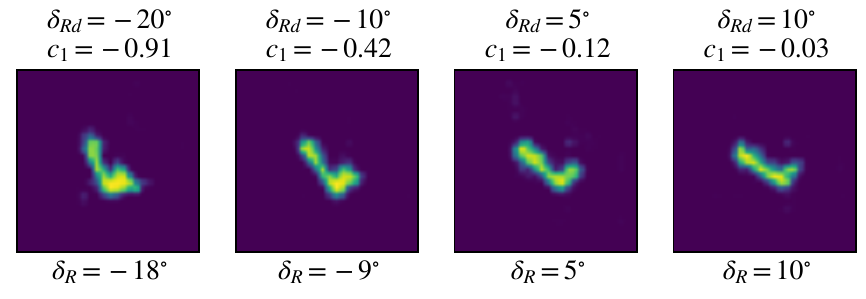}}		
	\caption{The results for the estimated and modeled rotation angle for the SAR images synthesized by the InfoGAN trained with real SAR images. The rotation angles in SAR image as a function of the latent code, $c_1$, measured by cross-correlation (black dots) and the estimated values with a linear model (green line) (top-left).  The rotation angles in SAR image as a function of the latent code, $c_1$, measured by cross-correlation (black dots) and the estimated values with a nonlinear model (yellow line) (top-right). The synthesized SAR images using $c_1$ calculated by (\ref{inversetanh}) for four desired rotation angles, $\delta_{Rd}=-20^{\circ}, \ -10^{\circ}, \ 5^{\circ},  \ 10^{\circ}$ (bottom row). The estimated rotations of the synthesized SAR images, $\delta_R(k)$ are calculated using (\ref{eq:rotation_corr}). }
	\label{RealDataImage}
\end{figure}

\begin{figure}[tpb]	
	\centering
	{\includegraphics{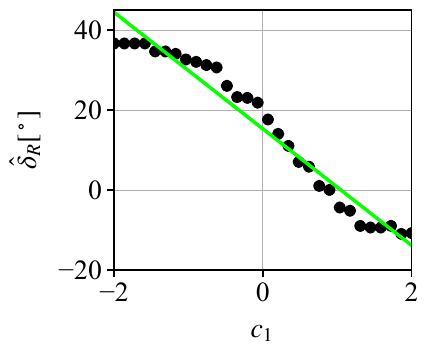}}
	{\includegraphics{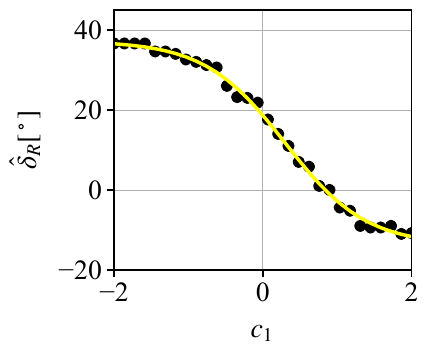}}
	{\includegraphics{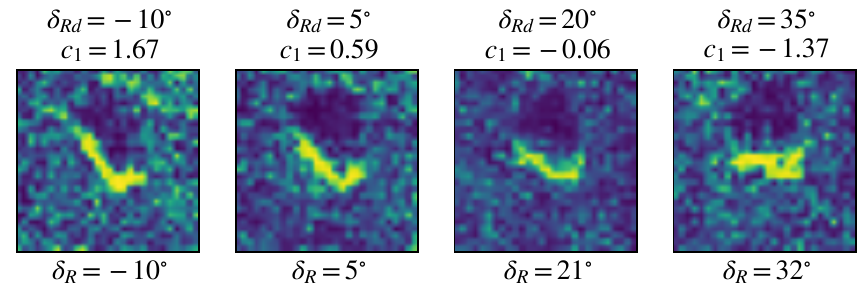}}		
	\caption{The results for the estimated and modeled rotation angle for the SAR images synthesized by the InfoGAN trained with real SAR images (not removing background). The organization of this figure is the same as Fig.~\ref{RealDataImage}.}
	\label{RealDataImagebackground}
\end{figure}

\begin{figure}[tbp]
	\centering
	{\includegraphics{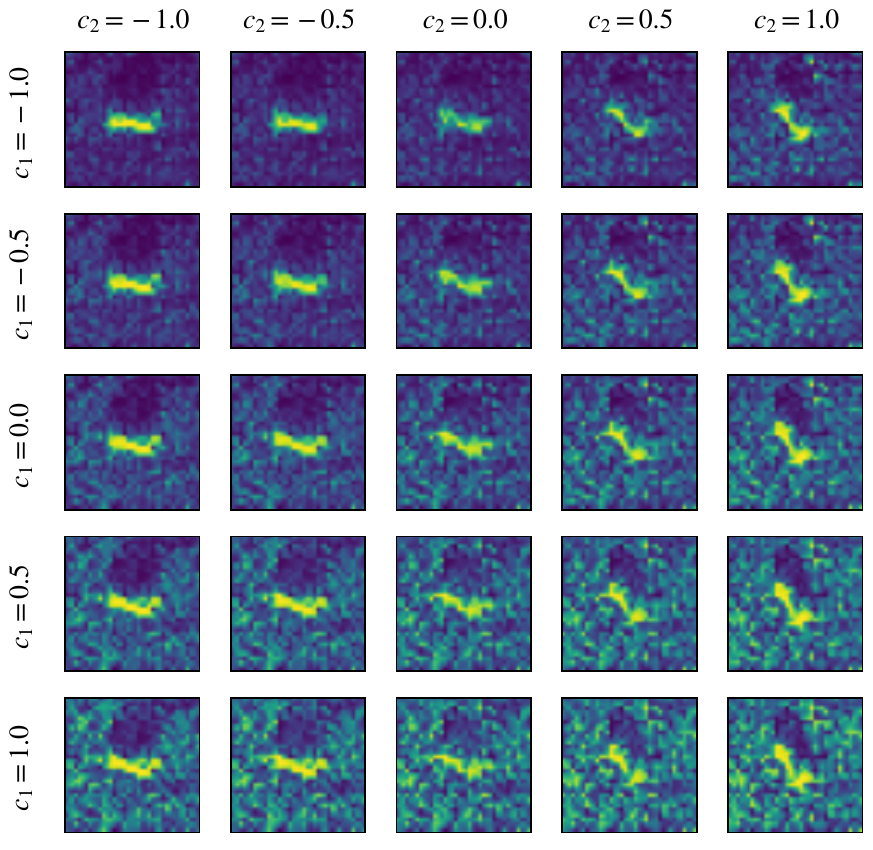}}
	\caption{The synthesized SAR images (with background). Two latent codes are used.}
	\label{RealDataImagebackground2C}
\end{figure}


The real-measured dataset is MSTAR with SAR images of ground stationary targets released by the MSTAR program supported by the Defense Advanced Research Projects Agency (DARPA) of the United States \colcit{\cite{MSTAR}}. The MSTAR dataset includes $2536$ SAR images for training and  $2636$ for testing with $10$ classes of vehicles. 
Different from the manual rotation in simulated data and semi-simulated data, the real rotation angle for each training SAR image is unknown.   We firstly choose a reference image and use (\ref{eq:rotation_corr}) to estimate the rotation of the rest images with respect to the reference one. As in the simulated example, the rotation is here accompanied by changes in intensity, resulting in changes of shape and a possible small mismatch between the (\ref{eq:rotation_corr}) and rotation angle. 
We have chosen $60$ images of 2S1 (self-propelled artillery) with rotation angles (with respect to one called reference SAR image) from  $[-34^\circ$, $44^\circ]$. The images are downsampled to the size of $28 \times 28$ pixels.


After the InfoGAN is trained with only $c_1$ activated, the same experiments as for simulated SAR images are conducted, as shown in Fig.~\ref{RealDataImage}. We can see that the latent code $c_1$, after the training process, is associated with the SAR image rotation. The modeling of the rotation angle and the latent code has been performed using the linear and nonlinear model, Fig.~\ref{RealDataImage} (top row). While the linear model is simple, the nonlinear model fits the data better. Finally, the model was used to synthesize new SAR images for a given desired rotation angle, $\delta_{Rd}$. The obtained images are shown in the bottom row of Fig.~\ref{RealDataImage} for four desired angles. The estimated rotation angles, $\hat{\delta}_{R}$ of the SAR images synthesized with $c_1$ calculated by (\ref{inversetanh}), is given in this panel, as well, and we can see that it is close to the desired ones,  $\delta_{Rd}$. 

\subsection{SAR Images with Background}
Furthermore, we conduct the same experiments with real SAR images, but now not removing the background, and the results are similar to the previous experiment, as shown in Fig.~\ref{RealDataImagebackground}, where the measured and modeled rotation angle is shown (with respect to the reference SAR image). Four synthesized SAR images with desired rotation, controlled by the latent code values, are given in Fig.~\ref{RealDataImagebackground}(bottom).  The experiment with the included background was repeated with two latent codes in the InfoGAN. Some synthesized SAR images are shown in Fig.~\ref{RealDataImagebackground2C}. As it can be seen from this figure, the latent code $c_1$ controls the rotation, while the latent code $c_2$, in this case, takes control over the background intensity. Thus, if we want to get images with suppressed background, we can use high values of $c_2$.

	
\section{Conclusions}\label{sec:Conclusion}

This article sheds some light on the relation between properties of synthesized SAR images and latent codes in InfoGAN, providing an analytical interpretation of this relation. The experiments are carried out with four datasets: simulated images, real objects from SAR image with simulated properties, SAR images with suppressed background, and SAR images with background. 
In the first experimental setup, the results demonstrate that the relation between a single latent code and one property matches a sigmoid function. In the second case, the results show that quadratic terms in the argument are required to cater to more complex relations when two latent codes are considered. The third and fourth experimental setups further demonstrate such a conclusion is applicable to real SAR images.
Therefore, it is possible to synthesize SAR images of these properties by manipulating latent codes according to such relation interpreted by our proposed method.

\bibliographystyle{IEEEtran}

\bibliography{References}

\end{document}